# On the Applicability of Zero-Shot Cross-Lingual Transfer Learning for Sentiment Classification in Distant Language Pairs

Andre Rusli    Makoto Shishido
Graduate School of Advanced Science and Technology
Tokyo Denki University
{20udc91@ms., shishido@mail.}dendai.ac.jp

## Abstract

*This research explores the applicability of cross-lingual transfer learning from English to Japanese and Indonesian using the XLM-R pre-trained model. The results are compared with several previous works, either by models using a similar zero-shot approach or a fully-supervised approach, to provide an overview of the zero-shot transfer learning approach's capability using XLM-R in comparison with existing models. Our models achieve the best result in one Japanese dataset and comparable results in other datasets in Japanese and Indonesian languages without being trained using the target language. Furthermore, the results suggest that it is possible to train a multi-lingual model, instead of one model for each language, and achieve promising results.*

## 1 Introduction

In recent years, the pace of newly proposed methods and tools for natural language processing (NLP) using deep learning approaches are increasing rapidly. After BERT [1], a massive pre-trained language model based on Transformers [2], was released, it only took a couple of years for the deep learning and NLP community to propose improvements and other methods that are shown to be better than the last, pushing various boundaries in the NLP field. While high-resource languages have achieved great successes in various tasks, other languages with limited data and computational resources are still left behind.

Cross-lingual transfer learning, where a high-resource language is used to train a downstream task model to improve the model's performance in another target language, shows a promising potential to tackle this setback. Many works of research have experimented and shown the potential benefit of using cross-lingual transfer in several NLP tasks, such as machine translation [3] [4], and named entity recognition [5] [6]. Recently, cross-lingual transfer learning has become an essential tool for improving performance in various downstream NLP tasks. This is also thanks to the recent advancements of massively multi-lingual Transformers pre-trained models, including mBERT [1], XLM [7], and the most recent one being XLM-RoBERTa (XLM-R) [8].

Even though the multi-lingual field has grown a lot, there are still some challenges. For example, previous works have reported that the quality of unsupervised cross-lingual word embedding is susceptible to the choice of language pairs and the comparability of the monolingual data [9], thus limiting the performance when the source and target language have different linguistic structures (e.g., English and Japanese) [10]. XLM-R is trained using 100 languages globally and achieved SOTA results in various multi-lingual NLP tasks such as the XNLI, Named Entity Recognition, Cross-lingual question answering, and the GLUE benchmark [8]. However, it reports only the performance evaluation for some of the languages used during pretraining such as English, French, Swahili, and Urdu.

This research explores the XLM-R base pre-trained model's capability for cross-lingual transfer learning encompassing English, Japanese, and Indonesian languages. We compare the performance with models from previous works, evaluate zero-shot transfer learning capability, and fine-tune mono- and multi-lingual models in a supervised manner for comparison purposes. This paper shows that both fine-tuning and zero-shot transfer learning from English to Japanese and Indonesian for a downstream task; in this case, binary sentiment classification, using XLM-R yields promising results. All experiments are conducted using Google Colaboratory. We also report the full specifications and hyperparameters used in our experiments for reproducibility and provide an overview of the applicability of using XLM-R for zero-shot transfer learning within a limited amount of data and computational resources.

## 2 Related Works

As a massively multi-lingual Transformers (MMT) model, XLM-R [8] is a robustly trained RoBERTa, exposed to a much larger multi-lingual corpus than mBERT. It is trained on the CommonCrawl-100 data of 100 languages. There are 88 languages in the intersection of XLM-R's and mBERT's corpora; for some languages (e.g., Kiswahili), XLM-R's monolingual data are several orders of magnitude larger than with mBERT. There are many methods for performing cross-lingual transfer based on MMTs, some of which are fine-tuning [11] and zero-shot transfer learning [12] [6]. The common thread is that data in a high-resource source language can be used to improve performance on a low-resource target language.

Even though XLM-R is pre-trained using 100 languages, investigations regarding the applicability of XLM-R for downstream tasks in some languages with less resource than English, such as Japanese and Indonesian, with thorough experiments and reproducible results are still limited. Several works have tried to implement cross-lingual transfer learning using several Japanese and Indonesian text classification models.

For Japanese, previous works have shown the capability of XLM-R for Japanese for dependency parsing [13] and named entity recognition [14], but no thorough comparison for cross-lingual transfer learning (fine-tuned and zero-shot) for sentiment classification are provided. In the Multi-lingual Amazon Review Corpus [15] in which Japanese is one of the languages of the corpus, the authors provided baseline sentiment classification performance using mBERT, but performance using XLM-R has not been reported. For Indonesian, previous works have shown the capability of using BERT and XLM-R for various tasks including sentiment classification [16] [17], however, the results are mainly focused on building powerful monolingual models for Indonesian.

Furthermore, unlike English, Japanese texts contain no whitespace and there are various ways to split sentences into words, with each split could end in a different meaning and nuance. In order to tackle this problem, a recent work proposed a language-independent subword tokenizer and detokenizer designed for neural-based text processing, named SentencePiece [18]. Its performance is shown to be effective for various tasks involving language pairs with different character sets, such as English-Japanese neural machine translation [18] and sentiment analysis in Japanese [19]. It is also utilized by state-of-the-art cross-lingual models such as XLM-R, which is the focus of our current research.

## 3 Experimental Setup

### 3.1 Dataset

We gathered binary sentiment datasets from several sources and put shorthand nicknames on each dataset to be addressed in the following sections.

1. AmazonEN: English Amazon product review sentiment dataset from The Multi-lingual Amazon Reviews Corpus [15]. We use 160,000 data for fine-tuning. 4,000 data for evaluation.
2. AmazonJA: Japanese Amazon product review sentiment dataset from The Multi-lingual Amazon Reviews Corpus [15]We use 160,000 data for fine-tuning. 4,000 data for evaluation.
3. RakutenJA: Japanese Rakuten product review binary sentiment dataset from [20]. We use 400,000 data for evaluation.
4. IndolemID: Indonesian Twitter and hotel review sentiment dataset from IndoLEM dataset [17]. We use 5,048 data for evaluation.
5. SmsaID: Indonesian multi-platform review sentiment dataset from SmSA dataset [21]. We use 1,129 data for evaluation.

For each dataset, we use the review body/text as the input and the sentiment (0 for negative and 1 for positive) as the classification label.

### 3.2 Experimental Setup

In this experiment, we use the free version of Google Colab with GPU for all our experiments. Due to the dynamic GPU allocation by Google Colab, two GPU types are used in our experiment: Tesla T4 and Tesla

P100-PCIE-16GB. Review sentiments gathered from AmazonEN and AmazonJA are rated as a 5-star rating. Following the original paper's practice (Keung et al., 2020), we converted the 1- and 2-stars rating as negative the 4- and 5-stars rating as positive reviews, we omit the 3-stars rating. Indonesian review data gathered from SmsaID initially contains three classes, which are positive, negative, and neutral; similarly, we omit the neutral class in this paper.

Furthermore, to provide an equal comparison of our models' performance, we use similar metrics used in the sources of each dataset. Specifically, we use the error percentage for AmazonEN, AmazonJA, RakutenJA, and the macro-averaged F1-score for SmsaID and IndolemID. Additionally, we report the hyper-parameters and the time needed for fine-tuning the models to provide a general overview of the applicability and resource needed by the readers to reproduce the results of our experiments. We divide the experiments into two scenarios:

1. Fine-tuned supervised learning
   Fine-tune the XLM-RoBERTa$_{BASE}$ pre-trained model using AmazonEN, AmazonJA, and the combination of English and Japanese Amazon reviews (AmazonENJA), then evaluate the models in monolingual and multi-lingual settings.
2. Zero-shot transfer learning
   Use the fine-tuned model using AmazonEN to evaluate zero-shot cross-lingual transfer learning capability in AmazonJA, RakutenJA, SmsaID, and IndolemID datasets.

## 4 Results and Analysis

### 4.1 Fine-tuned Supervised Learning

We fine-tuned the xlm-roberta-base pre-trained model from the HuggingFace transformers library, three times, using AmazonEN, AmazonJA, and AmazonENJA. For AmazonEN and AmazonJA, our final models are fine-tuned using the linear scheduler with warmup, 4 epochs, batch size=32, optimizer=AdamW, and learning rate=2e-5. For AmazonENJA, the final model uses the same above parameters but only with 2 epochs. Table 1 shows the GPU and averaged elapsed time for each epoch in the fine-tuning process.

**Table 1.** Fine-tuning specifications and elapsed time

| Fine-tuning Source Data | GPU | Epoch(s) | Average elapsed time per epoch |
|---|---|---|---|
| *AmazonEN* | Tesla T4 | 4 | 33 minutes 5 seconds |
| *AmazonJA* | Tesla P100-PCIE-16GB | 4 | 17 minutes 31 seconds |
| *AmazonENJA* | Tesla P100-PCIE-16GB | 2 | 35 minutes 57 seconds |

After fine-tuning three models in the previous step, we evaluate each of the fine-tuned models for supervised learning performance on the exact language from which the model is fine-tuned. We calculate the error percentage for the model prediction on test data and compare the results with a baseline model trained using mBERT [15], as displayed in Table 2. It can be seen that XLM-R outperforms mBERT in all three supervised models for binary sentiment classification in English and Japanese. There is no comparison from the baseline model for the multi-lingual model fine-tuned using AmazonEN and AmazonJA. However, it can be seen that it is possible to have a single bi-lingual model for both languages, eliminating the need to set up multiple models for every language.

**Table 2.** Error percentage of the fully-supervised evaluation on the Multi-lingual Amazon Review Corpus. Results using mBERT are obtained from [15].

| Model | EN-only | JA-only | EN&JA |
|---|---|---|---|
| mBERT | 8.8 | 11.1 | - |
| XLM-R$_{BASE}$ | **7.35** | **7.25** | **7.19** |

### 4.2 Zero-shot Transfer Learning

In this scenario, we use the fine-tuned models from the previous scenario to evaluate the applicability of zero-shot cross-lingual transfer learning from one language to the others. Table 3 shows the results (error percentage, lower is better) of the experiments conducted in this scenario using the multi-lingual Amazon and Japanese Rakuten data. Additionally, Table 4 reports the results (F-1 score, higher is better) using the multi-platform Indonesian sentiment datasets.

**Table 3.** Error percentage of zero-shot cross-lingual transfer learning using XLM-R$_{BASE}$ in comparison to a zero-shot mBERT from English data [15] and Japanese data [19]

| Model | *AmazonJA* | *RakutenJA* |
|---|---|---|
| Zero-shot mBERT | 19.04 | - |
| Fully-supervised ULMFiT | - | 4.45 |
| XLM-R$_{BASE}$ w/ *AmazonEN* | **11.12** | **13.09** |
| XLM-R$_{BASE}$ w/ *AmazonENJA* | 7.05 | 8.51 |

On the Japanese product review from the Amazon dataset, in Table 3, our model achieves a better error percentage of 11.12. It is almost 8 points better than the original baseline model [15], which is also evaluated using zero-shot cross-lingual transfer learning from English to Japanese, using mBERT. On another Japanese dataset with more evaluation data (400,000 reviews), the Rakuten dataset, our model achieves a 13.09 error percentage with zero-shot from English. Furthermore, it achieves a much better score of 8.51 error percentage if we add a substantial 160,000 review data from AmazonJA when fine-tuning the model. Although they are from different platforms, product review data shares similar patterns. This result is still far from the SOTA result of the 4.45 error percentage achieved by previous work [19], in which the model is trained in a fully-supervised monolingual setting using BERT.

For the Indonesian sentiment datasets, as also described in Table 4, we use two datasets with comparable results from different sources for evaluation purposes. Macro-averaged F1 score is used to evaluate the Indonesian datasets to follow previous works for comparison purposes. We experimented with zero-shot cross-lingual transfer learning using two models for classifying the Indonesian datasets; one is trained only with 160,000 English Amazon reviews, and another one containing English and Japanese Amazon reviews. In both Indonesian datasets, we can see a pattern of more data leads to better performance. Similar to prior research results [8], the model trained with multilingual data, in our case, Japanese and English review data, performs better. The XLM-R$_{BASE}$ w/ AmazonENJA model achieves a 73.31 F1-score on the IndolemID dataset, outperforming a previous model [17], trained with mBERT in a fully-supervised monolingual setting. Moreover, the same model achieves a better F1-score of 88 on the SmsaID dataset, outperforming another mBERT model [16] trained in a fully-supervised monolingual setting. For both datasets, our model performance is still worse when compared to the SOTA model's result, as shown in Table 3 and Table 4.

**Table 4**. Macro-averaged F1-score of zero-shot cross-lingual transfer learning using XLM-R$_{BASE}$ for Indonesian (IndoLEM [17], and SmSA [16]).

| Model | IndolemID | SmsaID |
|---|---|---|
| Fully-supervised BERT | 84.13 | 92.72 |
| Fully-supervised mBERT | 76.58 | 84.14 |
| XLM-R$_{BASE}$ w/ *AmazonEN* | 72.19 | 86.77 |
| XLM-R$_{BASE}$ w/ *AmazonENJA* | 73.31 | 87.99 |

Based on the results above, it is essential to note that the models used to compare RakutenJA, IndolemID, and SmsaID are trained in a fully-supervised approach using the same language with the target language. In contrast, our model, which is trained using AmazonEN, has never seen Japanese product reviews, and our models trained with AmazonEN and AmazonENJA have never seen Indonesian review texts. Furthermore, it is also essential to know that the models compared in the previous tables are trained mainly by much bigger architectures with more epochs, which means training them will need much more computational resources and time. Our current experiments show the applicability of zero-shot cross-lingual transfer learning with less computational costs, which yields promising results.

## 5 Conclusion and Future Work

This paper reports the results of experiments focusing on evaluating the applicability of cross-lingual transfer learning using the XLM-R pre-trained model. We then compare the results with previous works to provide an overview of the zero-shot approach's capability using XLM-R to use a multi-lingual model for bilingual data instead of one model for one language. Based on the results, zero-shot cross-lingual transfer learning yields promising results using XLM-R. All experiments are performed using the free version of Google Colab. The models achieve the best result in one dataset and shows the applicability of cross-lingual transfer learning, considering that the models have not seen languages in the target dataset, it can outperform SOTA results in other datasets trained in a fully-supervised approach. Future research steps include experimenting with more hyperparameters, evaluating other potential methods such as few-shot transfer learning, which has been proven to be useful to improve performance by adding just a few annotated data [13] and meta-learning [22].

# References

[1] J. Devlin, M.-W. Chang, K. Lee and K. Toutanova, "BERT: Pre-training of Deep Bidirectional Transformers for Language Understanding," in *NAACL*, Minneapolis, Minnesota, 2019.

[2] A. Vaswani, N. Shazeer, N. Parmar, J. Uszkoreit, L. Jones, A. N. Gomez, L. Kaiser and I. Polosukhin, "Attention Is All You Need," in *arXiv preprint arXiv:1706.03762.*, 2017.

[3] T. Q. Nguyen and D. Chiang, "Transfer Learning across Low-Resource, Related Languages for Neural Machine Translation," in *Eighth International Joint Conference on Natural Language Processing*, Taipei, 2017.

[4] G. Neubig and J. Hu, "Rapid Adaptation of Neural Machine Translation to New Languages," in *Conference on Empirical Methods in Natural Language Processing*, Brussels, 2018.

[5] S. Mayhew, C.-T. Tsai and D. Roth, "Cheap translation for cross-lingual named entity recognition," in *Conference on Empirical Methods in Natural Language Processing*, Copenhagen, 2017.

[6] J. Xie, Z. Yang, G. Neubig, N. A. Smith and J. Carbonell, "Neural Cross-Lingual Named Entity Recognition with Minimal Resources," in *Conference on Empirical Methods in Natural Language Process- ing*, Brussels, 2018.

[7] G. Lample and A. Conneau, "Cross-lingual Language Model Pretraining," in *arXiv preprint arXiv:1901.07291*, 2019.

[8] A. Conneau, K. Khandelwal, N. Goyal, V. Chaudhary, G. Wenzek, F. Guzman, E. Grave, M. Ott, L. Zettlemoyer and V. Stoyanov, "Unsupervised Cross-lingual Representation Learning at Scale," in *arXiv preprint arXiv:1911.02116*, 2020.

[9] A. Søgaard, S. Ruder and I. Vulić, "On the limitations of unsupervised bilingual dictionary induction," 2018.

[10] X. Chen, A. H. Awadallah, H. Hassan, W. Wang and C. Cardie, "Multi-Source Cross-Lingual Model Transfer: Learning What to Share," 2018.

[11] B. Zoph, D. Yuret, J. May and K. Knight, "Transfer Learning for Low-Resource Neural Machine Translation," in *Conference on Empirical Methods in Natural Language Processing*, Austin, Texas , 2016.

[12] W. U. Ahmad, Z. Zhang, X. Ma, E. Hovy, K.-W. Chang and N. Peng, "On Difficulties of Cross-Lingual Transfer with Order Differences: A Case Study on Dependency Parsing," 2019.

[13] A. Lauscher, V. Ravishankar, I. Vulić and G. Glavaš, "From Zero to Hero: On the Limitations of Zero-Shot Cross-Lingual Transfer with Multilingual Transformers," in *Conference on Empirical Methods in Natural Language Processing (EMNLP)*, Online, 2020.

[14] J. Pfeiffer, I. Vulić, I. Gurevych and S. Ruder, "MAD-X: An Adapter-Based Framework for Multi-Task Cross-Lingual Transfer," 2020.

[15] P. Keung, Y. Lu, G. Szarvas and N. A. Smith, "The Multilingual Amazon Reviews Corpus," 2020.

[16] B. Willie, K. Vincentio, G. I. Winata, S. Cahyawijaya, X. Li, Z. Y. Lim, S. Soleman, R. Mahendra, P. Fung, S. Bahar and A. Purwarianti, "IndoNLU: Benchmark and Resources for Evaluating Indonesian Natural Language Understanding," in *1st Conference of the Asia-Pacific Chapter of the Association for Computational Linguistics and the 10th International Joint Conference on Natural Language Processing*, Suzhou, China, 2020.

[17] F. Koto, A. Rahimi, J. H. Lau and T. Baldwin, "IndoLEM and IndoBERT: A Benchmark Dataset and Pre-trained Language Model for Indonesian NLP," in *28th International Conference on Computational Linguistics*, Barcelona, Spain (Online) , 2020.

[18] T. Kudo and J. Richardson, "SentencePiece: A simple and language independent subword tokenizer and detokenizer for Neural Text Processing," in *Proceedings of the 2018 Conference on Empirical Methods in Natural Language Processing: System Demonstrations*, Brussels, 2018.

[19] E. Bataa and J. Wu, "An Investigation of Transfer Learning-Based Sentiment Analysis in Japanese," in *57th Annual Meeting of the Association for Computational Linguistics*, Florence, 2019.

[20] X. Zhang and Y. LeCun, "Which Encoding is the Best for Text Classification in Chinese, English, Japanese and Korean?," arXiv:1708.02657v2 [cs.CL], 2017.

[21] A. Purwarianti and I. A. P. A. Crisdayanti, "Improving Bi-LSTM Performance for Indonesian Sentiment Analysis Using Paragraph Vector," in *International Conference of Advanced Informatics: Concepts, Theory and Applications (ICAICTA)*, Yogyakarta, Indonesia, 2019.

[22] F. Nooralahzadeh, G. Bekoulis, J. Bjerva and I. Augenstein, "Zero-Shot Cross-Lingual Transfer with Meta Learning," in *Conference on Empirical Methods in Natural Language Processing (EMNLP)*, Online, 2020.